\PassOptionsToPackage{numbers, compress}{natbib}
\documentclass{article}
\usepackage[preprint]{neurips_2025}

\usepackage[utf8]{inputenc}
\usepackage[T1]{fontenc}
\usepackage{hyperref}
\usepackage{url}
\usepackage{booktabs}
\usepackage{amsmath, amssymb, amsfonts, bm}
\usepackage{nicefrac}
\usepackage{microtype}
\usepackage{xcolor}
\usepackage{graphicx}
\usepackage{subcaption}
\usepackage{wrapfig}
\usepackage{multirow}
\usepackage{makecell}
\usepackage{enumitem}
\usepackage{array}
\usepackage{colortbl}
\usepackage{pgfplots}
\pgfplotsset{compat=1.18}

\usepackage{algorithm}
\usepackage{algpseudocode}

\usepackage{tikz}

\usepackage{pifont}

\title{ACE: Exploring Activation Cosine Similarity and Variance for Accurate and Calibration-Efficient \\ LLM Pruning}

\author{%
  Zhendong Mi$^{1}$, Zhenglun Kong$^{2}$, Geng Yuan$^{3}$, Shaoyi Huang$^{1}$\\
  $^{1}$Stevens Institute of Technology \quad
  $^{2}$Harvard University \quad
  $^{3}$University of Georgia \\
  \texttt{\small \{zmi2,shuang59\}@stevens.edu}, 
  \texttt{\small Zhenglun\_Kong@hms.harvard.edu}, 
  \texttt{\small geng.yuan@uga.edu}
}

\begin{document}


\maketitle

\begin{abstract}
  With the rapid expansion of large language models (LLMs), the demand for memory and computational resources has grown significantly. Recent advances in LLM pruning aim to reduce the size and computational cost of these models. However, existing methods often suffer from either suboptimal pruning performance or low time efficiency during the pruning process.
In this work, we propose an efficient and effective pruning method that simultaneously achieves high pruning performance and fast pruning speed with calibration efficiency. Our approach introduces two key innovations:
(1) activation cosine similarity loss guided pruning metric: a novel method that considers the angular deviation of the output activation from dense model and pruned model.
(2) activation variance-guided pruning metric: a new metric that allows for better semantic information distinction preservation in the output
activations after pruning, enabling pruning effectiveness with smaller input sequence lengths.
These two components can be readily combined to enhance LLMs pruning in both accuracy and calibration efficiency. Experimental results show that we can achieve up to  18\% decrease of perplexity and up to 63\% less pruning time
on prevalent LLMs, such as LLaMA, LLaMA-2, and OPT. 
\end{abstract}

\section{Introduction}

Recently, large language models (LLMs) have emerged as a prominent area of investigation, demonstrating exceptional capabilities through extensive parameterization across various tasks, such as language understanding~\cite{devlin2018bert}, text generation~\cite{brown2020language, touvron2023llama}, question answering~\cite{rajpurkar2016squad, lewis2020retrieval}, dialogue~\cite{roller2021recipes}, and code generation~\cite{chen2021evaluating}, etc.
While the increasing scale of LLMs has yield substantial accuracy inprovements, the advancement necessitates a compromise in memory consumption and inference latency \cite{devlin2019bert, touvron2023llama, agarwal2023llm}.
%
For instance, deploying a LLaMA-65B model requires at least four A100-40GB GPUs, with the time-to-first-token (TTFT) exceeding 100 milliseconds~\cite{yang2025wanda++}, highlighting the significant limitations of practical deployment in resource-constrained environments.
To mitigate the computational bottlenecks, various models compression techniques have been proposed,
such as quantization \cite{bai2020binarybert, frantar2022spdy, xiao2023smoothquant, lin2024awq}, 
pruning \cite{wolff1992optimal, lecun1989optimal, mocanu2018scalable, sun2023simple, frantar2023sparsegpt}, weight decomposition \cite{hsu2022language, yang2024loretta}, etc.
Among them, LLMs post-training pruning~\cite{frantar2023sparsegpt, sun2023simple} has garnered particular attention due to their ability in
applying sparsity constraints to pre-trained LLMs without requiring computationally expensive retraining procedures,
thus avoiding the prohibitive memory overhead.


Although existing LLMs post-training pruning methods~\cite{sun2023simple, frantar2023sparsegpt} have demonstrated potential in compressing model size with reduced memory overhead and negligible accuracy loss across diverse tasks, these approaches typically employ a well-designed weight importance evaluation metric with numerical magnitudes of weights and activations to identify important weight elements that should be 
preserved during pruning. 
In this work, \textit{we identify two promising yet unexplored opportunities in designing the importance evaluation metrics via exploring the semantic information inherent in the input activation feature space}:
1) \textbf{Angular deviation should be minimized during pruning to preserve semantic integrity of LLMs:} most existing pruning criteria focus on
removing weight elements that exhibit
minimal sensitivity to the loss function,
emphasizing numerical magnitudes of weights and activations. 
However, these approaches demonstrate limited consideration for
the angular deviation of word representations within the embedding space, which is a factor that can fundamentally 
impact the semantic integrity of LLMs. Moreover, established research~\cite{ethayarajh2019contextual, li2020degeneration, mikolov2013efficient} has shown that both
semantic similarity and model performance are heavily dependent on the relative orientation of token embeddings. Consequently, maintaining directional consistency through the minimization of angular deviation is a critical requirement for preserving
semantic integrity in pruned models;
2) \textbf{For equal-valued weights, those with lower input activation variance more effectively maintain token-level semantic distinctions:} previous studies~\cite{ethayarajh2019contextual, gao2021simcse} have shown that reduced token-level variation can result in semantic collapse and performance degradation across both classification and generation tasks. 
However, existing works fail to consider the variance of input activation features, a critical factor in preserving semantic distinctions. 
Our empirical observation reveals that input activation variance directly correlates with semantic differentiation between tokens.
Specifically, when comparing two same-valued weights,
the weight associated with higher input activation produces reduced output activation differentiation across distinct tokens, thereby diminishing semantic distinctions. Consequently, such weights should be assigned lower importance scores compared to those exhibiting smaller variance, as they contribute less effectively to maintaining semantic diversity.


In this work, we propose ACE, which explores \underline{a}ctivation cosine similarity and variance for accurate and \underline{c}alibration-\underline{e}fficient LLMs pruning.
First, we analyze that if one weight element is pruned, then there is an angular deviation for output activations between the original dense model and the pruned model. 
Based on the analysis, we introduce the activation cosine similarity loss guided pruning metric (CosP), which considers
both angular deviation and value loss of the output activation. Second, inspired by the role of  input activation feature variance, we design an activation variance-guided weight pruning metric (VarP), which incorporates a variance-based perturbation term and allows for better semantic information distinction preservation in the output activations after pruning. Moreover, we provide a theoretically analysis on the calibration efficiency of our approach and show that our method can achieve high accuracy even when applied with reduced sequence lengths for calibration data, demonstrating the potential of the proposed method in practical deployment scenarios with limited calibration data.
%
Furthermore, VarP maintains (or even surpasses) the performance of full sequence length pruning baselines with fewer input sequence length and reduced pruning time, highlighting its effectiveness and calibration efficiency. 
We summarize our contributions as follows:

\begin{itemize}[leftmargin=*, noitemsep, topsep=0pt]
    \item We propose an activation cosine similarity loss-guided pruning metric, which incorporates the angular deviation of output activation to better preserve the semantic integrity during pruning.
    \item We propose the activation variance-guided  pruning metric, which includes the variance of input activation to avoid the diminish of distinction between different tokens during pruning.
    \item  We theoretically analyze our proposed method can achieve calibration efficiency. Moreover, the experimental results demonstrate that our approach can achieve
    high accuracy on the pruned models with less input calibration sequence length and reduced pruning time. 
    \item We conduct extensive experiments 
    on various LLMs, such as LLaMa, LLaMa-2 and OPT models. Experimental results show that our method can outperform the baselines for both unstructured sparsity and N:M sparsity settings. For example, 
    the combined method of CosP and VarP only takes 40\%-50\% of the pruning time to perform 2:4 semi-structured pruning on LLaMA-30B, while even obtaining a 0.15–0.2 reduction in perplexity compared with Wanda and RIA, respectively.
    
\end{itemize}

\section{Background and Related Work}

\textbf{Network Pruning for Neural Networks.} Pruning is a well-established technique for neural network compression \cite{han2015deep,liu2018rethinking}. 
Both unstructured pruning \cite{han2015learning, frankle2019lottery} and structured pruning \cite{liu2017learning, molchanov2019importance} were extensively explored for model compression and acceleration. 
Unstructured pruning approaches typically identify and remove individual weights based on criteria such as magnitude \cite{han2015learning} or gradient information \cite{lee2018snip}. While achieving high theoretical sparsity, these methods often require specialized or good hardware to realize actual speedups. 
Structured pruning methods, meanwhile, focus on removing entire structural components such as neurons, filters, or channels \cite{li2017pruning, liu2017learning}. 
Beyond unstructured pruning, structured sparsity patterns have attracted significant attention due to their hardware acceleration capabilities. In particular, the N:M semi-structured sparsity format \cite{mishra2021accelerating} has gained prominence, where N out of every M consecutive weights are retained (i.e., N:M sparsity). This pattern is directly accelerated by NVIDIA's Ampere and later GPU architectures through specialized hardware support. N:M semi-structured pruning has emerged as a prominent research focus, primarily due to its compatibility with modern high-performance GPU architectures, which allows for efficient deployment and substantial acceleration during inference \cite{sun2023simple, frantar2023optimal, zhang2024plug}.


\textbf{Post-Training Pruning for Large Language Models.}
%
Unlike training-aware sparsification~\cite{gale2019state}, which iteratively prunes and fine-tunes the model during training, post-training pruning (PTP) operates directly on pretrained checkpoints. This makes PTP appealing for scenarios with limited training access or budget. However, designing effective pruning metrics remains a key challenge. Early works such as Wanda~\cite{sun2023simple} rely on the element-wise product of weight magnitudes and input activations to estimate importance. Others, like RIA~\cite{zhang2024plug}, incorporate relative importance between input and output channels to mitigate the problem of channel collapse. Pruner-zero \cite{dong2024pruner} leverages evolutionary search to adaptively discover layer-wise metrics, while SparseGPT~\cite{frantar2023sparsegpt} formulates pruning as a local reconstruction problem inspired by second-order approximations.

However, there are some problems with these PTP methods. First, works like Wanda \cite{sun2023wanda} and RIA \cite{zhang2024plug} generally focus on the magnitude of output activations, ignoring the angular change of these activations in the embedding space, resulting in the direction change of output activations in the embedding space. Second, some of the studies like OPTISHEAR \cite{liu2025optishear} try to use a genetic algorithm to search for layer-wise pruning ratios that better preserve model performance, which will introduce significant computational overhead in the pruning process though it can achieve better performance. Finally, current works about the PTP method primarily focus on the enhancement of the element of the weights \cite{damadi2021compression, dong2019hawq, sun2023simple, frantar2023optimal}. Nevertheless, there are few studies to explore the statistical information of input activations to further improve the pruning performance.

\section{Proposed Method} \label{method}
\label{gen_inst}


In this section, we will describe the motivation and our proposed ACE, an accurate and calibration efficient pruning approach of Large Language Models (LLMs). 
Firstly, we propose a novel activation cosine similarity loss-guided pruning metric which evaluates the importance of weights based on its impact on the output activation angular deviation in vector space.
%
Secondly, we propose an activation variance-guided pruning metric 
which incorporates weights and the variance of input activations into the metric, aiming to maintain the relative distances between token representations in the embedding space during pruning,
thus higher accuracy of the pruned model. Thirdly, we combine the two inspirations and form our final proposed weight pruning method. Moreover, we theoretically analyze the calibration efficiency of our method.
%








\subsection{Activation Cosine Similarity Loss Guided Pruning Metric (CosP)} \label{cos}




\begin{wrapfigure}[14]{r}{0.5\textwidth}
\vspace{-0.28in}
\begin{center}
	\includegraphics[width = 0.5\textwidth]{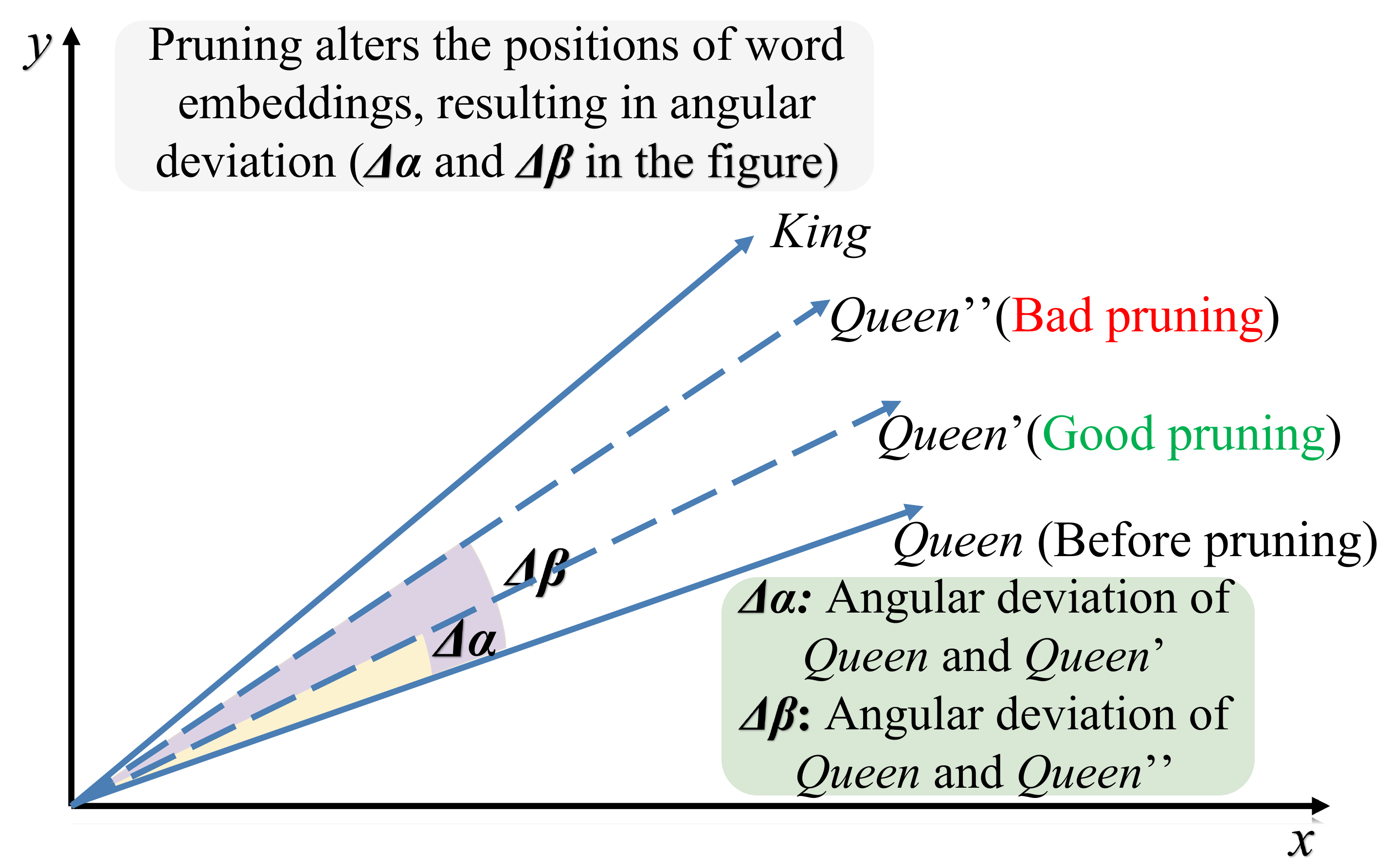} 
	\caption{Example of angular deviation before and after pruning}
	\label{fig:vector_space}
\end{center}
\end{wrapfigure}

\textbf{Motivation.} 
%
In Natural Language Processing (NLP), words are conventionally embedded into vector space. And semantically similar words exhibit approximate vector lengths and vector positions, where the latter is characterized by the small angular distances between them, resulting in cosine similarity value approaching 1~\cite{mikolov2013distributed, pennington2014glove}.
While pruning is utilized as a popular technique for model compression to improve efficiency, the resultant activation vectors exhibit non-trivial positional perturbations within the embedding space. As shown in Figure~\ref{fig:vector_space}, after pruning, the vector representations show deviations from the original position. Specifically, \textit{Queen} has shifted to \textit{Queen'} with a small angular deviation labeled as $\Delta \alpha$ after good pruning and has shifted to \textit{Queen''} with a large angular deviation labeled as $\Delta \beta$ after bad pruning. Bad pruning will potentially lead to the semantic error of a word.
Given that cosine similarity functions as a fundamental metric for quantifying semantic relationships between word embeddings in LLMs, one natural approach to evaluate the importance of the weight element is to incorporate the relative approximation loss of this metric into the pruning criteria for preserving semantic integrity.

\textbf{CosP Design.} We aim to use relative cosine-similarity loss as the metric to approximate the angular deviation when an individual weight element 
is pruned (set as zero). 
Consider a weight matrix $\textbf{W}$ of shape ($d_{out}$, $d_{in}$). 
In this layer of the neural network, we use $\textbf{X}$ with a shape of ($N$, $d_{in}$) to denote the input activation, where $N$ is the sequence length.
We use the cosine-similarity loss to represent the change in cosine similarity of the output sequence vector when an element of the weight matrix is pruned, which is formulated as: 
\begin{equation}
{L_{{cos}_{ij}}} = 1 - \cos \hat{\theta}_{ij} \approx -\frac{a_i \Delta a_i}{\|\bm{a}\|^2}=\frac{a_i\cdot \textbf{W}_{ij}x_j}{\|\bm{a}\|^2}
\end{equation}
where $\hat{\theta}_{ij}$ is angular change of the pruned model output activation (remove $\textbf{W}_{ij}$) and dense model output activation, 
$1 - cos\hat{\theta}_{ij}$ is the corresponding cosine similarity loss. A smaller $\hat{\theta}$ means smaller angular deviation and larger $\cos \hat{\theta}$ (i.e., closer to 1), therefore smaller loss in cosine similarity.
We use $\bm{\textbf{\textit{a}}}^\top=(a_1,a_2,...,a_i,..., a_{d_{out}})$ to denote the output activation for an input token ($\bm{\textbf{\textit{a}}}=\textbf{W}\bm{x}$). $\Delta a_i=-\textbf{W}_{ij}x_j$ denotes the loss caused by pruned weight element $\textbf{W}_{ij}$. $\bm{\textbf{\textit{x}}}^\top=(x_1,x_2,..., x_{d_{in}})$ represents an input vector.


Taking cosine similarity loss, weight value and output activation value into consideration, we formulate the proposed cosine similarity-guided pruning metric as follows:
\begin{equation}
\textbf{S}_{{cos}_{ij}} =
   |\textbf{W}_{ij}|\cdot ||\textbf{X}_{j}||_2 \cdot {L_{{cos}_{ij}}}
  =  |\textbf{W}_{ij}|\cdot ||\textbf{X}_{j}||_2 \cdot (1 - \cos \hat{\theta}_{ij})
\end{equation}  


We further derive the cosine similarity loss-guided importance score for the weight element $\textbf{W}_{ij}$ as:
\begin{equation}
\textbf{S}_{{cos}_{ij}} =
   |\textbf{W}_{ij}|\cdot ||\textbf{X}_{j}||_2 \cdot \frac{|\textbf{W}_{ij}|}{ \sqrt{\sum_{i = 1}^{d_{out}} \textbf{W}_{ij}^2/d_{out}}} \cdot \sqrt{\sum_{j = 1}^{d_{in}} \textbf{W}_{ij}^2}
\end{equation}
where $S_{{cos}_{ij}}$ denotes the relative cosine similarity pruning metric when the element $\textbf{W}_{ij}$ in the weight matrix $\textbf{W}$ is pruned, $\frac{|\textbf{W}_{ij}|}{ \sqrt{\sum_{i = 1}^{d_{out}} \textbf{W}_{ij}^2/d_{out}}} \cdot \sqrt{\sum_{j = 1}^{d_{in}} \textbf{W}_{ij}^2}$ approximates the relative cosine-similarity incurred by removing the element $\textbf{W}_{ij}$.
The detailed derivation can be found in the Appendix~\ref{appa}.

\subsection{Activation Variance-guided Weight Pruning Metric (VarP)} \label{var}


\begin{figure*}
    \centering
    \includegraphics[width=0.99\textwidth]{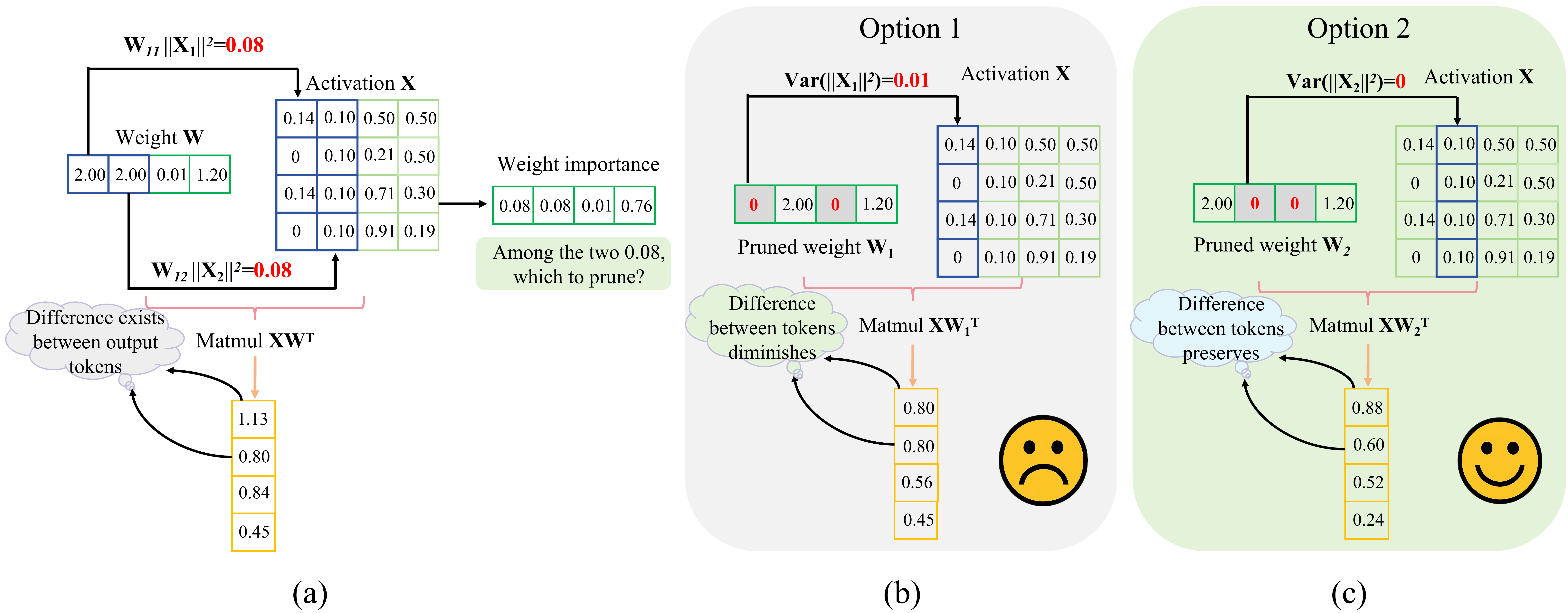}
    \caption{The motivating example of our proposed activation variance-guided  pruning metric}
    \vspace{-0.1in}
    \label{fig:main}
\end{figure*}

\textbf{Motivation.} Most existing LLM pruning methods rely on importance metrics computed through various formulations involving weights and activations.
However, the limitation arises when multiple elements share the same importance score. As illustrated in Figure 2(a), when two elements in the weight importance matrix have the same value (i.e., 0.08), 
it becomes difficult to determine which corresponding element to prune in the weight matrix $\textbf{W}$ to achieve a target sparsity of 50\%.
Two different pruning options exist based on the choice of pruning weight elements with the same importance score.
\textbf{Option 1} is to prune the first weight element with a corresponding larger variance (i.e., 0.01 > 0), resulting in a reduced difference between the first two elements in the output matmul compared to the original dense model,
as illustrated in Figure 2(b). In contrast, \textbf{Option 2} prunes the second element, which better preserves the output disparities between the first two elements, as shown in Figure 2(c), thereby maintaining closer alignment with the original distribution characteristics. 
For NLP tasks, preserving distinctions between tokens in the embedding space is crucial for maintaining semantic coherence and preventing the loss of token-level differences in model outputs \cite{ethayarajh2019contextual, li2020bert}.

\textbf{VarP Design.} Motivated by the above example, we propose our activation variance-guided weight importance score metric which incorporates the variance of input activation as follows:
\begin{equation}
\textbf{S}_{{var}_{ij}} =
|\textbf{W}_{ij}|\cdot (h(||\textbf{X}_{j}||_2) + \mathrm{Var}[||\textbf{X}_{j}||_2^2])
\label{eq:score_var}
\end{equation}


where $||\textbf{X}_{j}||_2$ is the $l_2$ norm of $j$th features
aggregated across $N$ different tokens, and $\mathrm{Var}[||\textbf{X}_{j}||_2^2]$ represents the variance of the squared values in the $j$-th column of the input activation.
$h( \cdot )$ is used to represent the transformation of the $l_2$ norm and serve as a factor in the product of $|\textbf{W}_{ij}|$ and  $h(||\textbf{X}_{j}||_2)$
to estimate the impact on 
the output when the weight element $\textbf{W}_{ij}$ is removed. 
Building upon $h(||\textbf{X}_{j}||_2)$ , we introduce an input variance-based perturbation term $\mathrm{Var}[||\textbf{X}_{j}||_2^2]$ to further determine which weight element should be pruned when the values of $|\textbf{W}_{ij}|\cdot h(||\textbf{X}_{j}||_2)$ are similar.
%
%
We define  $h(||\textbf{X}_{j}||_2)$ as follows to evaluate the importance of the $j$th feature in input activation $\textbf{X}$
\begin{equation}
h(||\textbf{X}_{j}||_2)= 
(\mathbb{E}[||\textbf{X}_{j}||_2^2])^2  + \mathbb{E}[||\textbf{X}_{j}||_2^2] +1
\label{eq:h_X}
\end{equation}
where $\mathbb{E}[||\textbf{X}_{j}||^2]$ represents the mean of the squared values in the $j$-th row of the input activation.
Combining Equation~\ref{eq:score_var} and Equation~\ref{eq:h_X}, we get our weight importance score metric as
\begin{equation}
\textbf{S}_{{var}_{ij}} =|\textbf{W}_{ij}|\cdot ((\mathbb{E}[||\textbf{X}_{j}||_2^2])^2 + \mathrm{Var}[||\textbf{X}_{j}||_2^2] + \mathbb{E}[||\textbf{X}_{j}||_2^2] +1)
\label{eq:score_var_2}
\end{equation}
Based on the formula linking variance and expectation, we have
\begin{equation}
\mathrm{Var}[X] = \mathbb{E}[X^2] - (\mathbb{E}[X])^2
\label{eq:var_X}
\end{equation}
Combing Equation~\ref{eq:score_var_2} and Equation~\ref{eq:var_X}, $\textbf{S}_{{var}_{ij}}$ can be further derived as follows
\begin{equation}
\textbf{S}_{{var}_{ij}} =|\textbf{W}_{ij}|\cdot (\mathbb{E}[||\textbf{X}_{j}||_2^4]  + \mathbb{E}[||\textbf{X}_{j}||_2^2] +1)
\end{equation}

Based on the fact that the input activations are normalized, implying that the corresponding activation values are  less than 1. Then, based on the power series expansion
\begin{equation}
 \mathbb{E}[||\textbf{X}_{j}||_2^4]  + \mathbb{E}[||\textbf{X}_{j}||_2^2] +1=\mathbb{E}[||\textbf{X}_{j}||_2^4 +||\textbf{X}_{j}||_2^2+1]\approx\mathbb{E}[\frac{1}{1-||\textbf{X}_{j}||_2^2}]
\end{equation}
We derive the importance score as
\begin{equation}
\textbf{S}_{{var}_{ij}} =|\textbf{W}_{ij}|\cdot \mathbb{E}[\frac{1}{1-||\textbf{X}_{j}||_2^2}]
\label{eq:score_var_final}
\end{equation}

\subsection{Our Final Weight Pruning Metric}

Take benefits from both factors ( i.e., cosine similarity and variance), 
we can combine the method proposed in Section \ref{cos} and Section \ref{var} to form our final weight pruning metric as follows

\begin{equation}
\textbf{S}_{{(cos+var)}_{ij}} =
   |\textbf{W}_{ij}|\cdot \mathbb{E}[\frac{1}{1-||\textbf{X}_{j}||_2^2}] \cdot \frac{|\textbf{W}_{ij}|}{ \sqrt{\sum_{i = 1}^{d_{out}} \textbf{W}_{ij}^2/d_{out}}} \cdot \sqrt{\sum_{j = 1}^{d_{in}} \textbf{W}_{ij}^2}
\end{equation}


\subsection{Calibration Data Efficiency Analysis}



SparseGPT~\cite{frantar2023sparsegpt} formulates LLMs post-training pruning as a layer-wise reconstruction problem, where for each layer, it aims to minimize the reconstruction error after pruning. Drawing inspiration from Optimal Brain Surgeon (OBS)~\cite{hassibi1993optimal}, SparseGPT~\cite{frantar2023sparsegpt} develops a pruning metric as follows
\begin{equation}
\textbf{S}_{ij} =  \frac{|\textbf{W}_{ij}|^2}{\operatorname{diag}\left( (\textbf{X}^T \textbf{X} + \lambda \textbf{I})^{-1} \right)}_{j},
\label{eq:13}
\end{equation}
where $\textbf{X}^T \textbf{X} + \lambda \textbf{I}$ represents the regularized Hessian matrix used in the layer-wise reconstruction problem
and $\lambda$ is used to prevent algorithm failure due to singular matrices thus ensuring the Hessian is always invertible.
Given the input activation as $\textbf{X}=(\textbf{X}_1,\textbf{X}_2,\dots,\textbf{X}_{d_{in}})$, we have 
\begin{equation}
    \frac{1}{\operatorname{diag}\left( (\textbf{X}^T \textbf{X} + \lambda \textbf{I})^{-1} \right)_j } = \frac{\lambda + ||\textbf{X}||^2}{\lambda + \\||\textbf{X}\\||^2 - \textbf{X}_j^2} \cdot \lambda
    \label{eq:14}
\end{equation}
Where \( \|\textbf{X}\|^2 = \sum_{i=1}^{d_{in}} \textbf{X}_i^2 \).
Wanda \cite{sun2023simple} uses a coarse formulation to approximate $\operatorname{diag}( (\textbf{X}^T \textbf{X} + \lambda I)^{-1})$ as follows
\begin{equation}
    \frac{1}{\operatorname{diag}\left( (\textbf{X}^T \textbf{X} + \lambda \textbf{I})^{-1} \right)_j } \approx \frac{1}{(\operatorname{diag} (\textbf{X}^T \textbf{X} + \lambda \textbf{I}))^{-1}_j}= \textbf{X}_j^2 +\lambda
    \label{eq:15}
\end{equation}
Taking the difference of our derivation (i.e., Equation~\ref{eq:14}) and Wanda's approximation (i.e., Equation~\ref{eq:15}), we have
\begin{equation}
\mathit{diff} = |\frac{\lambda + ||\textbf{X}||^2}{\lambda + \\||\textbf{X}\\||^2 - \textbf{X}_j^2} \cdot \lambda- (\textbf{X}_j^2 + \lambda)| \approx \frac{||\textbf{X}||^2\textbf{X}_j^2}{\lambda+||\textbf{X}||^2}
\label{eq:16}
\end{equation}
Suppose the input sequence length is denoted as $N$, we can further derive $diff$ as follows
(Detailed derivation can be found at Appendix \ref{appb2})
\begin{equation}
\mathit{diff}=\frac{1}{N+1}\mathbb{E}[||\textbf{X}_{j}||^2]
\label{eq:17}
\end{equation}
Equation~\ref{eq:17} shows an inverse relationship between the sequence length $N$ and $\mathit{diff}$. Specifically, as $N$ decreases, $\mathit{diff}$ increases monotonically, which demonstrates that our proposed method yields reduced reconstruction error and improved accuracy, particularly in scenarios with smaller input sequence length.
This theoretical finding suggests that our approach exhibits calibration data efficiency.
The detailed derivation can be found in Appendix \ref{appb1} and \ref{appb2}.

\section{Experiments}
\subsection{Experimental Setup}
\textbf{Models and Evaluations.} We work with
LLaMa 7B-65B~\cite{touvron2023llama}, LLaMA2 7B-13B~\cite{touvron2023llama2}, and OPT 350M-13B~\cite{zhang2022opt} to evaluate our proposed method.
All model checkpoints used in our experiments are obtained from the HuggingFace Transformers library to ensure reproducibility. 
%
For fair comparison, we employ uniform pruning across all linear layers while preserving the
embeddings and the head as dense~\cite{sun2023simple,zhang2024plug}. 
We evaluate the proposed method 
on zero-shot tasks and language modeling task. For zero-shot evaluation, we evaluate on seven benchmark tasks from EleutherAI LM Harness~\cite{gao2021framework} following existing work~\cite{sun2023simple} on LLaMa models.
For the language modeling task, we measure the perplexity of the three model families on the WikiText~\cite{merity2016pointer}. All experiments are conducted on a server with 8 NVIDIA A100 GPUs, each  with 40GB memory.

 \textbf{Baselines.}  For both unstructured pruning and N:M semi-structured pruning, we compare our proposed method with: 
 1) Wanda~\cite{sun2023simple}, which employs a pruning criterion based on the product of each weight matrix element and the sum of the norms of the corresponding input activations, and 2) RIA~\cite{zhang2024plug}, which modifies Wanda's pruning criterion by replacing each weight value with the sum of its relative contribution within its row and column, and uses this as the pruning score. 

 \textbf{Calibration Data.} For fair comparison with baselines, we take 128 samples from the C4 dataset~\cite{raffel2020exploring} for all models.
 Max context length size is used for both unstructured pruning and N:M semi-structured pruning for Wanda and RIA.

 \subsection{Language Modeling}

 \subsubsection{Evaluation on CosP }

\begin{table}[ht]
\centering
\small
\captionsetup{justification=raggedright,singlelinecheck=false}
\caption{Perplexity 
of OPT, LLaMA and LLaMA-2 on Wikitext2. In the experiment, we all use full sequence length for pruning (2048 for LLaMa and OPT, 4096 for LLaMa-2)} 
\vspace{0.08in}
\resizebox{1.0\linewidth}{!}{\begin{tabular}{lccccccc|cccc|cc}
\toprule
\multirow{2}{*}{Method}  & \multirow{2}{*}{Sparsity} & \multicolumn{6}{c}{OPT}   & \multicolumn{4}{|c}{LLaMA} & \multicolumn{2}{|c}{LLaMA-2} \\
\cmidrule(lr){3-8} \cmidrule(lr){9-12} \cmidrule(lr){13-14}
  & & 350M & 1.3B & 2.7B & 6.7B & 13B & 30B  & 7B & 13B & 30B & 65B & 7B & 13B  \\
\midrule
Dense &  0\% &22.03 & 14.62 & 12.47 & 10.86 & 10.12 & 9.55  & 5.68 & 5.09 & 4.77 & 3.56 & 5.12 & 4.57  \\
\midrule
\multirow{3}{*}{Wanda} 
&  50\% & 36.24 & 18.40 & 14.22 & 11.98 & 11.92 & 10.03  & 7.26 & 6.15 & 5.25 & 4.60 & 6.46 & 5.56  \\
&  2:4 & 114.57 & 28.15 & 21.27 & 15.91 & \textbf{15.53} & 13.47  & 11.53 & 9.60 & 6.89 & 6.24 & 11.34 & 8.35  \\
&  4:8 & 58.95 & 22.20 & 16.78 & 13.55 & \textbf{13.38} & 10.87   & 8.56 & 7.40 & 5.98 & 5.30 & 8.09 & 6.52  \\
\midrule
\multirow{3}{*}{RIA} 
&  50\% & 36.90 & 18.17 & 14.30 & 11.82 & 12.05 & 9.92   & 7.12 & 6.08 & \textbf{5.08} & \textbf{4.42} & 6.81 & 5.83  \\
&  2:4 & 114.89 & 27.43 & 21.69 & 15.75 & 15.88 & 13.38  & 11.10 & 8.97 & 6.74 & 6.00 & 10.58 & 7.85  \\
&  4:8 & 59.68 & 21.60 & 16.82 & 13.44 & 13.60 & 10.80  & 8.32 & 7.15 & 5.79 & 5.10 & 7.85 & 6.30 \\
\midrule
\multirow{3}{*}{CosP (Ours)} 
&  50\% & \textbf{36.40}& \textbf{18.15} & \textbf{14.21} & \textbf{11.77} & \textbf{11.92} &\textbf{9.91}   & \textbf{7.10} & \textbf{6.05} & 5.14 & 4.44 & \textbf{6.32} & \textbf{5.45}  \\
&  2:4 &\textbf{108.29} & \textbf{26.66} & \textbf{20.87} & \textbf{15.38} & 15.75 & \textbf{13.12} & \textbf{11.07} & \textbf{8.94} & \textbf{6.72} & \textbf{5.98} & \textbf{10.41} & \textbf{7.77}  \\
&  4:8 & \textbf{57.71}& \textbf{21.31} & \textbf{16.53} & \textbf{13.23} & \textbf{13.38} & \textbf{10.77}  & \textbf{8.28} & \textbf{7.11} & \textbf{5.79} & \textbf{5.10} & \textbf{7.70} & \textbf{6.28}  \\
\bottomrule
\end{tabular}}

\vspace{-0.15in}
\label{tab:opt-pruning}
\end{table}


We evaluate the proposed CosP and report the perplexity of pruned OPT model and LLaMa model in Table \ref{tab:opt-pruning}. Our findings are summarized as follows:

1) On the OPT and LLaMa models, our method consistently outperforms both Wanda and RIA baselines, especially on OPT models. For instance, on the OPT-1.3B model, CosP with unstructured pruning can achieve a PPL approximately 0.3 lower than that of Wanda. 

2) Our approach demonstrates even greater effectiveness under N:M semi-structured pruning compared to unstructured pruning on both LLaMa models and OPT models, yielding more pronounced performance compared to both Wanda and RIA. For example, on the OPT-350M model, our CosP 2:4 semi-structured pruning achieves a PPL of 108.29, nearly 6 points lower than RIA's 114.89. Similarly, on the OPT-1.3B model, our CosP 4:8 semi-structured pruning yields a PPL of 26.66, outperforming Wanda’s 28.15 by approximately 1.5.

\subsubsection{Evaluation on VarP and Final Pruning Metric}

We conduct experiments to evaluate our proposed VarP approach, including the perplexity on Wikitext2 and runtime. In Tables \ref{tab:opt-pruning-VarP} and Table \ref{tab:llama-pruning-VarP}, "FL" indicates that full sequence lengths is incorporated for pruning according to Wanda (e.g. 2048 for LLaMa and OPT, 4096 for LLaMa-2), “16” indicates that only 16 sequences are sampled from each calibration input for pruning, while all other settings remain identical to those used in Wanda.

\begin{table}[ht]
\vspace{-4mm}  
\centering
\small
\caption{PPL and pruning time (s) of OPT model series on Wikitext2}
\vspace{0.08in}
\resizebox{\textwidth}{!}{
\begin{tabular}{lcccccccc}
\toprule
\multirow{3}{*}{Method} & \multirow{3}{*}{Seq\_Len} & \multirow{3}{*}{Sparsity} & \multicolumn{6}{c}{OPT} \\
\cmidrule(lr){4-9}
& & & 350M & 1.3B & 2.7B & 6.7B & 13B & 30B \\
\cmidrule(lr){4-9}
& & & \multicolumn{1}{c}{PPL \textbar\ Time} & \multicolumn{1}{c}{PPL \textbar\ Time} & \multicolumn{1}{c}{PPL \textbar\ Time} & \multicolumn{1}{c}{PPL \textbar\ Time} & \multicolumn{1}{c}{PPL \textbar\ Time} & \multicolumn{1}{c}{PPL \textbar\ Time} \\
\midrule

\multirow{3}{*}{Wanda-FL} & \multirow{3}{*}{FL} & 50\% & 36.24 \textbar\ 31.4 & 18.40 \textbar\ 39.4 & 14.22 \textbar\ 49.3 & 11.98 \textbar\ 67.5 & 11.92 \textbar\ 98.8 & 10.03 \textbar\ 183.5 \\
& & 2:4 & 114.57 \textbar\ 38.9 & 28.15 \textbar\ 50.8 & 21.27 \textbar\ 69.4 & 15.91 \textbar\ 111.7 & 15.53 \textbar\ 160.7 & 13.47 \textbar\ 300.6 \\
& & 4:8 & 58.95 \textbar\ 35.0 & 22.20 \textbar\ 42.8 & 16.78 \textbar\ 58.1 & 13.55 \textbar\ 87.6 & 13.38 \textbar\ 129.8 & 10.87 \textbar\ 240.4 \\
\midrule

\multirow{3}{*}{Wanda-16} & \multirow{3}{*}{16} & 50\% & 42.91 \textbar\ 17.8 & 24.62 \textbar\ 17.8 & 17.62 \textbar\ 21.2 & 14.33 \textbar\ 25.7 & 12.34 \textbar\ 31.3 & 11.01 \textbar\ 45.2 \\
& & 2:4 & 136.12 \textbar\ 24.8 & 35.63 \textbar\ 34.4 & 25.69 \textbar\ 47.8 & 18.58 \textbar\ 65.8 & 15.34 \textbar\ 97.8 & 19.13 \textbar\ 171.3 \\
& & 4:8 & 66.70 \textbar\ 20.7 & 27.82 \textbar\ 25.9 & 20.55 \textbar\ 34.7 & 17.00 \textbar\ 45.2 & 13.28 \textbar\ 63.3 & 12.58 \textbar\ 102.3 \\
\midrule

\multirow{3}{*}{RIA-16} & \multirow{3}{*}{16} & 50\% & 40.09 \textbar\ 16.1 & 20.46 \textbar\ 19.8 & 15.68 \textbar\ 21.8 & 12.32 \textbar\ 24.4 & 11.70 \textbar\ 31.6 & 10.22 \textbar\ 44.8 \\
& & 2:4 & 141.18 \textbar\ 23.1 & 30.88 \textbar\ 32.4 & 23.89 \textbar\ 46.8 & 16.25 \textbar\ 63.8 & 15.10 \textbar\ 95.7 & 16.89 \textbar\ 161.5 \\
& & 4:8 & 67.48 \textbar\ 20.2 & 23.43 \textbar\ 24.5 & 18.24 \textbar\ 34.8 & 14.78 \textbar\ 45.5 & 12.95 \textbar\ 64.5 & 12.00 \textbar\ 101.9 \\
\midrule

\multirow{3}{*}{VarP-16 (Ours)} & \multirow{3}{*}{16} & 50\% & 37.78 \textbar\ 25.8 & 20.77 \textbar\ 27.1 & 15.02 \textbar\ 33.2 & 12.32 \textbar\ 37.2 & 11.64 \textbar\ 45.1 & 10.29 \textbar\ 63.7 \\
& & 2:4 & 103.80 \textbar\ 33.8 & 29.59 \textbar\ 44.1 & 24.45 \textbar\ 57.7 & 16.11 \textbar\ 78.4 & 13.95 \textbar\ 112.5 & 14.02 \textbar\ 178.5 \\
& & 4:8 & 58.13 \textbar\ 31.0 & 23.34 \textbar\ 34.8 & 17.61 \textbar\ 46.0 & 13.77 \textbar\ 58.2 & 12.33 \textbar\ 76.4 & 11.13 \textbar\ 120.3 \\
\midrule

\multirow{3}{*}{CosP+VarP-16 (Ours)} & \multirow{3}{*}{16} & 50\% & 37.04 \textbar\ 25.3 & 19.85 \textbar\ 27.2 & 14.72 \textbar\ 33.8 & 12.09 \textbar\ 37.2 & 11.47 \textbar\ 45.1 & 10.18 \textbar\ 63.7 \\
& & 2:4 & 115.43 \textbar\ 33.4 & 28.57 \textbar\ 42.8 & 23.50 \textbar\ 59.8 & 15.49 \textbar\ 77.9 & 14.41 \textbar\ 124.5 & 13.23 \textbar\ 177.3 \\
& & 4:8 & 58.12 \textbar\ 29.5 & 22.51 \textbar\ 35.1 & 17.60 \textbar\ 47.1 & 13.50 \textbar\ 57.7 & 12.29 \textbar\ 77.5 & 11.05 \textbar\ 118.6 \\
\bottomrule
\end{tabular}
}

\vspace{-0.1in}
\label{tab:opt-pruning-VarP}
\end{table}

\begin{table}[ht]
\vspace{-5mm} 
\centering
\small

\caption{PPL and pruning time (s) of LLaMA and LLaMA-2 model series on Wikitext2}
\vspace{0.08in}
\resizebox{\textwidth}{!}{
\begin{tabular}{lclcccccc}
\toprule
\multirow{2}{*}{Method} & \multirow{2}{*}{Seq\_Len} & \multirow{2}{*}{Sparsity} & \multicolumn{4}{c}{LLaMA} & \multicolumn{2}{c}{LLaMA-2} \\
\cmidrule(lr){4-7} \cmidrule(lr){8-9}
& & & 7B & 13B & 30B & 65B & 7B & 13B \\
\cmidrule(lr){4-7} \cmidrule(lr){8-9}
& & & \multicolumn{1}{c}{PPL \textbar\ Time} & \multicolumn{1}{c}{PPL \textbar\ Time}
    & \multicolumn{1}{c}{PPL \textbar\ Time} & \multicolumn{1}{c}{PPL \textbar\ Time}
    & \multicolumn{1}{c}{PPL \textbar\ Time} & \multicolumn{1}{c}{PPL \textbar\ Time} \\
\midrule

\multirow{3}{*}{Wanda-FL} & \multirow{3}{*}{FL} & 50\% & 7.26 \textbar\ 64.6 & 6.15 \textbar\ 39.4 & 5.25 \textbar\ 49.3 & 4.60 \textbar\ 67.6 & 6.46 \textbar\ 98.8 & 5.56 \textbar\ 183.5 \\
& & 2:4 & 11.53 \textbar\ 80.6 & 9.60 \textbar\ 121.6 & 6.89 \textbar\ 260.9 & 6.24 \textbar\ 463.7 & 11.34 \textbar\ 179.4 & 8.35 \textbar\ 267.7 \\
& & 4:8 & 8.56 \textbar\ 70.2 & 7.40 \textbar\ 104.2 & 5.98 \textbar\ 226.8 & 5.30 \textbar\ 410.3 & 8.09 \textbar\ 160.2 & 6.52 \textbar\ 234.6 \\
\midrule

\multirow{3}{*}{Wanda-16} & \multirow{3}{*}{16} & 50\% & 7.89 \textbar\ 23.7 & 6.60 \textbar\ 29.8 & 5.41 \textbar\ 40.0 & 4.46 \textbar\ 60.7 & 6.82 \textbar\ 22.7 & 5.88 \textbar\ 27.9 \\
& & 2:4 & 11.70 \textbar\ 41.1 & 9.35 \textbar\ 55.7 & 7.00 \textbar\ 98.3 & 6.15 \textbar\ 152.7 & 11.73 \textbar\ 42.7 & 8.35 \textbar\ 55.3 \\
& & 4:8 & 8.69 \textbar\ 31.0 & 7.37 \textbar\ 40.2 & 5.87 \textbar\ 65.9 & 5.16 \textbar\ 100.9 & 8.28 \textbar\ 31.3 & 6.48 \textbar\ 40.1 \\
\midrule

\multirow{3}{*}{RIA-16} & \multirow{3}{*}{16} & 50\% & 7.27 \textbar\ 33.2 & 6.11 \textbar\ 41.6 & 5.14 \textbar\ 66.5 & 4.40 \textbar\ 102.1 & 6.44 \textbar\ 33.4 & 5.53 \textbar\ 43.7 \\
& & 2:4 & 11.53 \textbar\ 72.9 & 8.83 \textbar\ 102.5 & 6.81 \textbar\ 195.6 & 6.03 \textbar\ 323.7 & 11.17 \textbar\ 74.0 & 7.90 \textbar\ 108.4 \\
& & 4:8 & 8.46 \textbar\ 51.5 & 7.14 \textbar\ 72.8 & 5.78 \textbar\ 131.7 & 5.08 \textbar\ 209.3 & 8.05 \textbar\ 53.0 & 6.33 \textbar\ 74.6 \\
\midrule

\multirow{3}{*}{VarP-16 (Ours)} & \multirow{3}{*}{16} & 50\% & 7.18 \textbar\ 29.9 & 6.12 \textbar\ 37.4 & 5.13 \textbar\ 53.1 & 4.41 \textbar\ 78.4 & 6.49 \textbar\ 29.2 & 5.47 \textbar\ 36.7 \\
& & 2:4 & 11.03 \textbar\ 48.3 & 8.61 \textbar\ 65.9 & 6.66 \textbar\ 110.6 & 5.72 \textbar\ 174.4 & 10.57 \textbar\ 48.2 & 7.42 \textbar\ 65.2 \\
& & 4:8 & 8.25 \textbar\ 38.1 & 7.03 \textbar\ 57.2 & 5.76 \textbar\ 80.8 & 4.97 \textbar\ 121.5 & 7.95 \textbar\ 38.4 & 6.16 \textbar\ 50.5 \\
\midrule

\multirow{3}{*}{CosP+VarP-16 (Ours)} & \multirow{3}{*}{16} & 50\% & 7.11 \textbar\ 30.2 & 6.06 \textbar\ 37.1 & 5.11 \textbar\ 54.3 & 4.43 \textbar\ 79.7 & 6.47 \textbar\ 29.5 & 5.46 \textbar\ 36.2 \\
& & 2:4 & 11.18 \textbar\ 48.8 & 8.53 \textbar\ 64.9 & 6.64 \textbar\ 110.5 & 5.82 \textbar\ 173.8 & 11.02 \textbar\ 49.2 & 7.50 \textbar\ 64.3 \\
& & 4:8 & 8.25 \textbar\ 39.8 & 6.95 \textbar\ 50.7 & 5.72 \textbar\ 79.5 & 5.01 \textbar\ 123.2 & 7.93 \textbar\ 38.3 & 6.16 \textbar\ 52.1 \\
\bottomrule
\end{tabular}
}

\vspace{-0.2in}
\label{tab:llama-pruning-VarP}
\end{table}

Table \ref{tab:opt-pruning-VarP} and Table \ref{tab:llama-pruning-VarP} report the perplexity results on the WikiText2 dataset for our proposed VarP method compared to the Wanda and RIA baselines across OPT and LLaMa models. The key findings are summarized as follows:

1) Compared to the full-sequence pruning variant of Wanda-FL, our VarP-16 method offers substantial gains in time efficiency on both LLaMa and OPT models. For example, on the OPT-6.7B model with 2:4 semi-structured pruning, VarP-16 completes pruning in just 78.4 seconds with a perplexity (PPL) of 16.11, while Wanda-FL requires 111.7 seconds for a similar PPL of 15.91—resulting in approximately a 30\% reduction in pruning time with minimal performance loss.

2) When compared to RIA-16, VarP-16 exhibits slightly higher latency but yields significantly better evaluation performance on OPT models, but on LLaMa models,our method can achieve both high accuracy and lower latency. On the LLaMA-7B model with 2:4 semi-structured pruning, RIA-16 takes 72.9 seconds and produces a PPL of 11.53. In comparison, VarP-16 completes the process in only 48.3 seconds and achieves a lower PPL of 11.03, representing a 0.5 reduction in perplexity.

3) While Wanda-16 is the fastest among the methods, it comes at the cost of degraded performance. For example, on the OPT-2.7B model with unstructured pruning, VarP-16 achieves a PPL of 20.77, outperforming Wanda-16’s PPL of 24.62.

4) Building upon VarP-16, we observe that integrating the CosP method into it
can lead to enhanced pruning performance. The improvements are especially pronounced in N:M semi-structured settings. On OPT-1.3B model with 2:4 pruning, CosP yields a further reduction of 1.0 in perplexity compared to VarP-16, demonstrating its effectiveness in preserving model quality under structured sparsity.

\subsection{Zero-shot Accuracy}

\begin{table}[ht]
\vspace{-5mm} 

\caption{Mean zero-shot accuracy (\%)\(\uparrow\) of LLaMA, LLaMA-2 and OPT models on 7 zero-shot tasks}
\vspace{0.08in}
\centering
\small
\resizebox{0.8\columnwidth}{!}{
\begin{tabular}{lclcccccc}
\toprule
\multirow{2}{*}{Method} & \multirow{2}{*}{Seq\_Len} & \multirow{2}{*}{Sparsity} & \multicolumn{2}{c}{LLaMA} & \multicolumn{2}{c}{LLaMA-2} & \multicolumn{2}{c}{OPT} \\
\cmidrule(lr){4-5} \cmidrule(lr){6-7} \cmidrule(lr){8-9}
& & & 7B & 13B & 7B & 13B & 6.7B & 13B \\
\midrule
Dense & - & 0\% & 59.99 & 62.59 & 59.71 & 63.03 & 51.58 & 52.60 \\
\midrule
\multirow{3}{*}{Wanda-FL} 
& \multirow{3}{*}{FL} & 50\% & 54.21 & \textbf{59.33} & \textbf{56.24} & 59.81 & 47.05 & \textbf{51.28} \\
& & 2:4 & 48.53 & \textbf{52.30} & \textbf{48.57} & 53.08 & 45.78 & 47.56 \\
& & 4:8 & \textbf{52.76} & \textbf{55.10} & \textbf{52.49} & \textbf{58.75} & 47.32 & 49.24 \\
\midrule
\multirow{3}{*}{Wanda-16} 
& \multirow{3}{*}{16} & 50\% & 53.74 & 57.67 & 55.10 & 59.94 & 48.24 & 50.04 \\
& & 2:4 & 47.83 & 51.52 & 46.98 & 53.59 & 45.36 & 47.14 \\
& & 4:8 & 50.88 & 54.26 & 51.07 & 57.15 & 46.83 & 48.58 \\
\midrule
\multirow{3}{*}{RIA-16} 
& \multirow{3}{*}{16} & 50\% & 55.07 & 57.28 & 54.43 & 59.39 & 48.55 & 50.52 \\
& & 2:4 & 48.41 & 52.11 & 47.24 & 53.57 & 46.01 & 47.44 \\
& & 4:8 & 51.18 & 54.52 & 51.54 & 56.97 & 47.31 & 48.71 \\
\midrule
\multirow{3}{*}{CosP+VarP-16 (Ours)} 
& \multirow{3}{*}{16} & 50\% & \textbf{55.11} & 57.31 & 54.87 & \textbf{60.11} & \textbf{49.42} & 50.52 \\
& & 2:4 & \textbf{49.37} & \textbf{52.30} & 47.89 & \textbf{54.06} & \textbf{46.46} & \textbf{48.10} \\
& & 4:8 & 52.05 & 54.97 & 51.69 & 57.40 & \textbf{48.33} & \textbf{49.82} \\
\bottomrule
\end{tabular}}

\vspace{-0.07in}
\label{tab:zero-shot-accuracy}
\end{table}

Table \ref{tab:zero-shot-accuracy} reports the average zero-shot accuracy across seven tasks for some of the LLaMa model family and OPT model family. We also provide the detailed evaluation accuracy on the models of the seven datasets in our Appendix~\ref{appc3}. Across both unstructured and structured sparsity settings, our method significantly outperforms the widely used Wanda-16 and RIA-16 pruning with few sequences pruning baseline, and some of the results show that our method with the input sequence length of only 16 even surpasses the full sequence Wanda-FL pruning method. As a representative example, on the OPT-6.7B model with 2:4 structured pruning, our method achieves a zero-shot accuracy of 46.46\%, surpassing Wanda-16 by 1.1\%, RIA-16 by 0.45\%, and Wanda-FL by 0.78\%, respectively. As no fine-tuning is applied, a performance gap remains between the sparse pruned models and their dense counterparts.  Notably, we also can find that large size pruned model somehow can perform equivalently with dense small size model, with only about 1\%-2\% gap. Interestingly, performance of LLaMa-2-13B pruned model can achieve the accuracy of 60.11\%, exceeding the performance of LLaMa-2-7B dense model.

\subsection{Results of Different Sparsity Settings}

This section reports pruning outcomes across multiple sparsity configurations, with detailed results provided in Table \ref{tab:different_sparsity}. Our main findings are outlined below:

1) Under lower sparsity levels such as 20\% and 40\%, the RIA pruning method occasionally demonstrates stronger performance on certain models, particularly within the OPT family, while yielding results comparable to ours on LLaMA models. However, as the sparsity increases to 50\% or 60\%, our method consistently outperforms both RIA and Wanda in terms of pruning effectiveness. The most significant improvement is observed on the OPT-13B model, where our method reduces perplexity by approximately 1.9 under a 60\% unstructured pruning setting.

2) Throughout these experiments under varying sparsity settings, we find that our method maintains performance on par with, and occasionally surpasses, the full sequence length Wanda-FL baseline, despite using much shorter input sequence lengths. As an illustrative case, on the LLaMA-13B model with 60\% unstructured pruning, our method achieves a perplexity of 8.16, outperforming Wanda-FL’s 8.75 by a margin of 0.6. In contrast, applying the Wanda method with shorter input sequence lengths results in a substantial degradation in pruning performance, falling considerably behind both our approach and RIA across multiple settings.

\begin{table}[ht]
\vspace{-5.0mm} 
\centering
\small

\caption{Perplexity of LLMs with different sparsity on Wikitext2}
\vspace{0.07in}

\resizebox{0.8\columnwidth}{!}{
\begin{tabular}{lclcccccccc}
\toprule
\multirow{2}{*}{Method} & \multirow{2}{*}{Seq\_Len} & \multirow{2}{*}{Sparsity} & \multicolumn{2}{c}{LLaMA} & \multicolumn{2}{c}{LLaMA-2} & \multicolumn{3}{c}{OPT} \\
\cmidrule(lr){4-5} \cmidrule(lr){6-7} \cmidrule(lr){8-10}
& & & 7B & 13B & 7B & 13B & 1.3B & 6.7B & 13B \\
\midrule
\multirow{3}{*}{Wanda-FL} 
& \multirow{3}{*}{FL} & 20\% & 5.81 & 5.13 & 5.22 & 4.68 & 14.69 & 10.62 & 10.06 \\
& & 40\% & 6.39 & 5.51 & 5.66 & 5.01 & \textbf{15.87} & \textbf{10.96} & 10.64 \\
& & 60\% & \textbf{10.69} & 8.75 & \textbf{10.04} & 7.93 & \textbf{26.53} & 15.21 & 15.94 \\
\midrule
\multirow{3}{*}{Wanda-16} 
& \multirow{3}{*}{16} & 20\% & 5.76 & 5.13 & 5.17 & 5.09 & 15.32 & 10.43 & 9.94 \\
& & 40\% & 6.48 & 5.64 & 5.75 & 5.09 & 18.57 & 12.13 & 10.66 \\
& & 60\% & 13.35 & 9.75 & 11.34 & 8.60 & 45.77 & 20.32 & 17.52 \\
\midrule
\multirow{3}{*}{RIA-16} 
& \multirow{3}{*}{16} & 20\% & 5.75 & 5.13 & \textbf{5.16} & 4.62 & \textbf{14.34} & \textbf{10.38} & \textbf{10.00} \\
& & 40\% & 6.29 & 5.48 & \textbf{5.62} & 4.94 & 16.36 & 10.98 & \textbf{10.48} \\
& & 60\% & 11.23 & 8.44 & 10.16 & 7.58 & 32.76 & 16.21 & 15.68 \\
\midrule
\multirow{3}{*}{CosP+VarP-16 (Ours)} 
& \multirow{3}{*}{16} & 20\% & \textbf{5.74} & \textbf{5.12} & \textbf{5.16} & \textbf{4.59} & 14.96 & 10.93 & 10.16 \\
& & 40\% & \textbf{6.23} & \textbf{5.47} & \textbf{5.62} & \textbf{4.92} & 16.96 & 11.24 & 10.58 \\
& & 60\% & 10.82 & \textbf{8.16} & 10.82 & \textbf{7.28} & 28.37 & \textbf{15.04} & \textbf{14.15} \\
\bottomrule
\end{tabular}
}

\vspace{-0.1in}
\label{tab:different_sparsity}
\end{table}


\vspace{-0.1in}
\section{Limitations} \label{limit}
\vspace{-0.1in}
We consider the following limitations of our work:
1) Our method is specifically designed for post-training pruning. Whether it can be extended to the quantization domain, or integrated with existing quantization techniques, remains an open question and a promising direction for future research.
2) The current scope of our study is limited to large language models. In future work, we aim to explore the applicability of our approach to a broader range of architectures, such as vision models and vision-language models.

\vspace{-0.1in}

\section{Conclusion} \label{conc}
\vspace{-0.1in}
This work introduces an efficient and lightweight post-training pruning technique tailored for large language models which incorporates two components: CosP and VarP. The former (i.e., CosP) introduces a novel pruning criterion by explicitly considering the impact of weight removal on the cosine similarity of output activations. While the latter (i.e., VarP), incorporates input activation variance into the pruning metric, achieving both pruning effectiveness and calibration efficiency.
Experimental results show that both CosP and VarP can achieve efficient pruning with better performance than existing baselines. Moreover, our proposed LLM pruning method (CosP + VarP) can achieve the highest performance in both accuracy and calibration efficiency via taking benefit from both strategies.

\bibliographystyle{unsrt}
\bibliography{neurips_2025}            

\newpage
\appendix
\section{Appendix: Activation Cosine Similarity Loss Guided Pruning Metric } \label{appa}

\textbf{\textit{Lemma}}: Consider a vector $\bm{\textbf{a}} \in \mathbb{R}^d$ with elements $\bm{\textbf{\textit{a}}}=(a_1,a_2,...,a_k,...,a_d)$. If a small perturbation $\delta{a_k}$ is applied on a single element $a_k$, the resultant perturbed vector is given by 
$\bm{\textbf{\textit{a}}}'=(a_1,a_2,...,a_k+\delta{a_k},...)$.
Under the assumption that $|\delta{a_k}| \ll |a_k|$, the approximate cosine loss induced by the perturbation can be estimated as $-\frac{a_k\cdot\delta{a_k}}{\|\bm{\textbf{\textit{a}}}\|^2}$.

\textbf{\textit{Proof}}: If a vector undergoes a small perturbation, the norm of the perturbed vector can be expressed as follows:
\begin{equation}
\|\bm{a}'\| = \sqrt{ \sum_{i \ne k} a_i^2 + (a_k + \Delta a_k)^2 }
\end{equation}
Applying a first-order Taylor expansion with respect to $\Delta a_k$, we obtain the following approximation:
\begin{equation}
\|\bm{a}'\| \approx \|\bm{a}\| + \frac{a_k \Delta a_k}{\|\bm{a}\|}
\end{equation}

Since $|\delta{a_k}| \ll |a_k|$, the perturbed vector $\bm{a}'$ is close to $\bm{a}$.
Assume both $\bm{a}$ and $\bm{a}'$ are normalized (or nearly aligned), the cosine similarity between them can be approximated as:
\begin{equation}
\cos \hat{\theta} \approx \frac{\bm{a}^\top \bm{a}'}{\|\bm{a}\| \cdot \|\bm{a}'\|}
\end{equation}

We substitute the above equation into the cosine expression and expand:
\begin{equation}
\cos \hat{\theta} \approx \frac{\|\bm{a}\|^2 + a_k \Delta a_k}{\|\bm{a}\| \left( \|\bm{a}\| + \frac{a_k \Delta a_k}{\|\bm{a}\|} \right)}
= 1 + \frac{a_k \Delta a_k}{\|\bm{a}\|^2} + \mathcal{O}((\Delta a_k)^2)
\end{equation}

Then, the cosine loss (which measures dissimilarity) becomes:
\begin{equation}
1 - \cos \hat{\theta} \approx -\frac{a_k \Delta a_k}{\|\bm{a}\|^2}
\label{eq21}
\end{equation}

The above equation shows that the cosine loss increases approximately linearly with the product of the perturbed component and the magnitude of the perturbation.

Consider a linear layer with the output dimension $\textbf{\textit{d}}_{out} = m$ and the input dimension $\textbf{\textit{d}}_{in} = n$ . Given an input activation 
vector \( \bm{\textbf{X}^\top} = (\textbf{X}_1, \textbf{X}_2, \dots, \textbf{X}_n) \) and the weight matrix \( \textbf{W}\in \mathbb{R}^{m \times n} \), the forward pass computation can be expressed as $\bm{y^\top} = \left( \sum_{j=1}^n \textbf{W}_{1j} \textbf{X}_j,\ \sum_{j=1}^n \textbf{W}_{2j} \textbf{X}_j,\ \dots,\ \sum_{j=1}^n \textbf{W}_{mj} \textbf{X}_j \right)$ (To simplify the analysis, we assume the input activation $\bm{\textbf{X}^\top}$ is represented as a column vector, which is right-multiplied by the weight matrix 
\( \textbf{W}\in \mathbb{R}^{m \times n} \)).
To analyze the impact of pruning individual weight elements, we focus on the contribution (i.e., relative cosine similarity loss) of a specific neuron to the overall network output.
Taking the first neuron as an example and we analyze the effect of pruning each of its associated weight connections \( \textbf{W}_{11}, \textbf{W}_{21}, \dots, \textbf{W}_{m1} \) to the output. 
Based on the approximation of angular loss in Equation \ref{eq21}, if we remove the element \( \textbf{W}_{t1} \) for \( t \in (1, 2, \dots, m) \), the corresponding approximate absolute cosine loss (CL) can be denoted as:
\begin{equation}
CL_{t1}=|1 - \cos \hat\theta_{t1}| \approx \frac{|\textbf{W}_{t1} \textbf{X}_{1}| \left|( \textbf{W}_{t1} \textbf{X}_{1} + \textbf{W}_{t2} \textbf{X}_{2} + \dots + \textbf{W}_{tn} x_{n} \right)|}{\|\bm{y}\|^2}
\end{equation}
Then, we can use the following formula to quantify the relative Cosine-similarity importance score ($L_{{cos}_{t1}}$) resulting from pruning a specific element \( \textbf{W}_{t1} \):
\begin{equation}
L_{{cos}_{t1}} = \frac{CL_{t1}}{\sqrt{\frac{CL_{11}^2 + CL_{21}^2 + \dots + CL_{m1}^2}{m}}}
=\frac{\sqrt{m}|\textbf{W}_{t1}||(\textbf{W}_{t1}\textbf{X}_1+\textbf{W}_{t2}\textbf{X}_2+\dots+\textbf{W}_{tn}\textbf{X}_n)|}{\sqrt{ \sum_{i = 1}^{m} \textbf{W}_{i1}^2 \left( \sum_{j = 1}^{n} \textbf{W}_{ij} \textbf{X}_j \right)^2 }}
\label{eq23}
\end{equation}
Based on the Hölder's inequality as follows:
\begin{equation}
\sum_{i=1}^{n} |a_i b_i| \leq \left( \sum_{i=1}^{n} |a_i| \right) \cdot \max_{1 \leq i \leq n} |b_i|
\end{equation}
After substituting the above inequality into the denominator of the Equation~\ref{eq23}, we obtain:
\begin{equation}
{\sqrt{ \sum_{i = 1}^{m} \textbf{W}_{i1}^2 \left( \sum_{j = 1}^{n} \textbf{W}_{ij} \textbf{X}_j \right)^2 }}\leq
\sqrt{(\sum_{i = 1}^{m} \textbf{W}_{i1}^2)\cdot \max_{1 \leq i \leq m}(\sum_{j = 1}^{n} \textbf{W}_{ij} \textbf{X}_j )^2}
\end{equation}
For simplicity, we denote:
\begin{equation}
\max_{1 \leq i \leq m}(\sum_{j = 1}^{n} \textbf{W}_{ij} \textbf{X}_j )^2 = (\sum_{j = 1}^{n} \textbf{W}_{kj} \textbf{X}_j )^2
\end{equation}
Consequently, we obtain:
\begin{equation}
L_{{cos}_{t1}} \geq
\frac{|\textbf{W}_{t1}|}{ \sqrt{\sum_{i = 1}^{m} \textbf{W}_{i1}^2/m}} \cdot \frac{|\sum_{j = 1}^{n} \textbf{W}_{tj}\textbf{X}_j|}{|\sum_{j = 1}^{n} \textbf{W}_{kj} \textbf{X}_j| }
\label{eq:27}
\end{equation}
Given that the input 
\textbf{X} is remains constant during the computation of elements within the same row of weight matrix, we can derive two important simplification: 1) The expression $|\sum_{j = 1}^{n} \textbf{W}_{kj} \textbf{X}_j|$ represents a fixed value for the given weight matrix, therefore this term can be safely omitted when designing the pruning metric. The constancy of this term across different pruning candidates within the same row ensures that it does not affect the relatve ranking of importance scores; 2) For comparing the impact of different weight elements, 
the expression $|\sum_{j = 1}^{n} \textbf{W}_{tj}\textbf{X}_j|$ can be 
approximated by the simpler norm-based metric$\sqrt{\sum_{j = 1}^{n} \textbf{W}_{tj}^2}$ for different values of $t$. We have the approximation since this term becomes invariant across pruning decisions within the same row when pruning is performed row-wise, thereby having no effect on the final pruning outcome.
Therefore, it can be treated as a fixed scaling factor and safely omitted.
Finally, we derive the expression for $L_{{cos}_{t1}}$ of $\textbf{W}_{t1}$ as:
\begin{equation}
L_{{cos}_{t1}} \approx
\frac{|\textbf{W}_{t1}|}{ \sqrt{\sum_{i = 1}^{m} \textbf{W}_{i1}^2/m}} \cdot \sqrt{\sum_{j = 1}^{m} \textbf{W}_{tj}^2}
\end{equation}
The derivation presented for the specific weight element $W_{t1}$ can be extended to all other weight elements. Based on the theoretical analysis, we integrate the cosine loss term $L_{\cos_{t1}}$ with existing magnitude-based importance measures. Specifically, we incorporate the Wanda-based scoring mechanism to capture complementary important factors.
\begin{equation}
\textbf{S}_{{cos}_{ij}} =
\frac{|\textbf{W}_{ij}|}{ \sqrt{\sum_{i = 1}^{m} \textbf{W}_{ij}^2/m}} \cdot \sqrt{\sum_{j = 1}^{m} \textbf{W}_{ij}^2}  \cdot |\textbf{W}_{ij}|\cdot ||\textbf{X}_{j}||_2
\end{equation}

\section{Appendix: Activation Variance-guided Weight Pruning Metric} \label{appb}

\subsection{Derivation of VarP Importance Metric} \label{appb1}

Given the loss function \( L(\textbf{W}) \), its Taylor expansion around the current weights \( \textbf{W} \) is:

\begin{equation}
L(\textbf{W} + \Delta \textbf{W}) \approx L(\textbf{W}) + \nabla L(\textbf{W})^T \Delta \textbf{W} + \frac{1}{2} \Delta \textbf{W}^T \textbf{H} \Delta \textbf{W}
\end{equation}

where:
- \( \nabla L(\textbf{W}) \) is the gradient,
- \( \textbf{H} = \nabla^2 L(\textbf{W}) \) is the Hessian (second derivative matrix).
To minimize the loss variation after pruning, we focus on the second-order term \( \frac{1}{2} \Delta \textbf{W}^T \textbf{H} \Delta \textbf{W} \). However, calculating the full Hessian is computationally expensive for large models. SparseGPT approximates the Hessian using the input data matrix \( \textbf{X} \). Specifically, for MSE loss or linearized networks, \( \textbf{H} \approx \textbf{X}^T \textbf{X} \) and \( \lambda \textbf{I} \) is added as a regularization term to stabilize the inverse. Thus, the approximate local Hessian becomes:

\begin{equation}
\textbf{H} \approx \textbf{X}^T \textbf{X} + \lambda \textbf{I}
\end{equation}

The pruning metric used in SparseGPT is:
\begin{equation}
\textbf{S}_{ij} = \left[ \frac{|\textbf{W}|^2}{\operatorname{diag}\left( (\textbf{X}^T \textbf{X} + \lambda \textbf{I})^{-1} \right)} \right]_{ij}
\label{eq32}
\end{equation}
which captures how important a weight is, normalized by the local curvature (second-order sensitivity) of the loss surface.
If \( \textbf{X} \in \mathbb{R}^{1 \times n} \) is an input:
- \( \textbf{X}^T \in \mathbb{R}^{n \times 1} \),
- \( \textbf{X}^T \textbf{X} \in \mathbb{R}^{n \times n} \), a rank-1 matrix \( \textbf{uv}^T \) where \( \textbf{u} = \textbf{v} = \textbf{X}^T \). Using the Sherman-Morrison formula, we have:
\begin{equation}
(\lambda \textbf{I} + \textbf{uv}^T)^{-1} = \frac{1}{\lambda}\textbf{I} - \frac{1}{\lambda^2} \frac{\textbf{uv}^T}{1 + \frac{\textbf{v}^T \textbf{u}}{\lambda}}
\end{equation}
Specifically, the \( i \)-th diagonal element is:
\begin{equation}
\left[ (\textbf{X}^T \textbf{X} + \lambda \textbf{I})^{-1} \right]_{ii} = \frac{1}{\lambda} - \frac{1}{\lambda^2} \cdot \frac{\textbf{X}_i^2}{1 + \frac{\|\textbf{X}\|^2}{\lambda}}
\end{equation}
where \( \|\textbf{X}\|^2 = \sum_{i=1}^n \textbf{X}_i^2 \). Then, we have :
\begin{equation}
\operatorname{diag}\left( (\textbf{X}^T \textbf{X} + \lambda \textbf{I})^{-1} \right)_j = \frac{1}{\lambda} - \frac{\textbf{X}_j^2}{\lambda^2 + \lambda \|\textbf{X}\|^2}
\end{equation}
Each diagonal element depends on the square of the corresponding feature in \( X \). Then, we have:
\begin{equation}
    \frac{1}{\operatorname{diag}\left( (\textbf{X}^T \textbf{X} + \lambda \textbf{I})^{-1} \right)_j } = \frac{\lambda + ||\textbf{X}||^2}{\lambda + \\||\textbf{X}\\||^2 - \textbf{X}_j^2} \cdot \lambda
    \label{eq36}
\end{equation}
Note that the input \textbf{X} have been layernormed, so \( \|\textbf{X}\|^2\) can be approximately viewed as a constant, we can denote it as $k$. The pruning metric can be simplified as:
\begin{equation}
\textbf{S}_{ij} =  \frac{|\textbf{W}_{ij}|^2}{\operatorname{diag}\left( (\textbf{X}^T \textbf{X} + \lambda \textbf{I})^{-1} \right)} _{j}=|\textbf{W}_{ij}|^2 \cdot  \frac{\lambda + k}{\lambda + k - \textbf{X}_j^2} \cdot \lambda
\end{equation}
Since $\lambda$ is a constant, we can omit it and we can get:
\begin{equation}
\textbf{S}_{ij} =  \frac{|\textbf{W}_{ij}|^2}{\operatorname{diag}\left( (\textbf{X}^T \textbf{X} + \lambda \textbf{I})^{-1} \right)} _{j}=|\textbf{W}_{ij}|^2 \cdot  \frac{1}{1 - \frac{\textbf{X}_j^2}{\lambda+k}} 
\end{equation}
where $\lambda+k$ is a constant as well and we can omit it. 
We can use more input to calculate their mean value of calibration data and get our final pruning metric:
\begin{equation}
\textbf{S}_{{var}_{ij}}=\mathbb{E}[\textbf{S}_{ij}] =|\textbf{W}_{ij}|^2 \cdot  \mathbb{E}[\frac{1}{1 - ||\textbf{X}_{j}||^2}] 
\end{equation}

\subsection{Efficiency Analysis} \label{appb2}

Now we analyze the efficiency of VarP.  Wanda assumes the denominator of  Equation \ref{eq32} can be approximated:
\begin{equation}
    \frac{1}{\operatorname{diag}\left( (\textbf{X}^T \textbf{X} + \lambda \textbf{I})^{-1} \right)_j } \approx \frac{1}{(\operatorname{diag} (\textbf{X}^T \textbf{X} + \lambda \textbf{I}))^{-1}_j}= \textbf{X}_j^2 +\lambda
    \label{eq40}
\end{equation}
Taking the difference of the right sides between Equation \ref{eq36} and Equation \ref{eq40}, we have ($||\textbf{X}||^2$ can be approximately replaced by $k$):
\begin{equation}
    |\frac{\lambda(\lambda+k)}{\lambda+k-\textbf{X}_j^2}- \textbf{X}_j^2 -\lambda|
    =|\frac{\lambda}{1-\frac{\textbf{X}_j^2}{\lambda+k}}-\textbf{X}_j^2-\lambda|
\end{equation}
Applying the power series expansion, we have:
\begin{equation}
   |\frac{\lambda}{1-\frac{\textbf{X}_j^2}{\lambda+k}}-\textbf{X}_j^2-\lambda|=|\lambda(1+\frac{\textbf{X}_j^2}{\lambda+k})-\textbf{X}_j^2-\lambda|=\frac{k\textbf{X}_j^2}{\lambda+k}
\end{equation}
If we have more sequences with length of $L$ as input, we can take the average of the $\textbf{X}_j^2$, then we can 
the dfference between VarP and Wanda's as:
\begin{equation}
diff=\frac{k}{\lambda+k}\frac{\textbf{X}_{j1}^2+\textbf{X}_{j2}^2+\dots+\textbf{X}_{jn}^2}{L}=\frac{k}{\lambda+k}\mathbb{E}[||\textbf{X}_{j}||^2]
\end{equation}
Since the $\lambda$ is a constant, we can make it equals to $nk$, then:
\begin{equation}
diff=\frac{k}{\lambda+k}\mathbb{E}[||\textbf{X}_{j}||^2]=\frac{1}{L+1}\mathbb{E}[||\textbf{X}_{j}||^2]
\end{equation}
We can see from the above Equation that if $L$ is small, the difference between these two methods will be large, which demonstrates the calibration efficiency of our proposed method.

\section{Appendix: More Experiment Results} \label{appc}

\subsection{Robustness Analysis} \label{appc1}

\vspace{-0.1in}

To evaluate the robustness and stability of our method, we conduct experiments across multiple calibration data sampling configurations 
We select three different random seeds, therefore three different calibration data subsets, and repeat the pruning process for each of them on LLaMA-7B model using Wikitext2 dataset. 
It's shown in Table \ref{tab:seed} that our method  exhibits better results across different random seeds.

\begin{table}[htbp]
\centering
\vspace{-0.1in}
\caption{Perplexity for pruned LLaMA-7B models with different random seeds on Wikitext2}
\small

\vspace{0.05in}
\begin{tabular}{llccc}
\toprule
\textbf{Model} & \textbf{Method} & \textbf{seed\#1} & \textbf{seed\#2} & \textbf{seed\#3}  \\
\midrule
\multirow{3}{*}{LLaMA-7B} 
  & Wanda-16  & 7.89 & 8.19 & 7.76  \\
  & RIA-16  & 7.27 & 7.28 & 7.26  \\
  & CosP+VarP-16 (Ours)      & \textbf{7.11} & \textbf{7.11} & \textbf{7.12} \\

\bottomrule
\end{tabular}

\label{tab:seed}
\end{table}

\vspace{-0.2in}

\subsection{Impact of Sequence Lengths} \label{appc2}

\vspace{-0.1in}

\begin{table}[htbp]
\centering
\vspace{-0.1in}
\caption{Perplexity of pruned LLaMA-7B model with different sequence lengths on Wikitext2}
\small
\vspace{0.05in}
\begin{tabular}{>{\centering\arraybackslash}m{1.8cm} >{\centering\arraybackslash}m{2.5cm} *{6}{c}}
\toprule
\multirow{2}{*}{\textbf{Model}} & \multirow{2}{*}{\textbf{Method}} & \multicolumn{6}{c}{\textbf{Sequence Lengths}} \\
\cmidrule(lr){3-8}
& & \textbf{16} & \textbf{32} & \textbf{128} & \textbf{512} & \textbf{1024} & \textbf{2048} \\
\midrule
\multirow{3}{*}{\centering LLaMA-7B} 
  & Wanda      &     7.89   &    7.52    &   7.28    &    7.27   &  7.27     &    7.26   \\
  & RIA        &     7.27   &    7.22    &    7.15   &   7.13    &   7.12    &    7.12   \\
  & CosP+VarP (Ours)  &    7.11    &   7.12     &   7.13    &  7.17     &    7.18   &    7.18   \\
\bottomrule
\end{tabular}
\label{tab:seq}
\end{table}

We present the impact of sequence lengths on the LLaMA-7B model for Wikitext2 under 50\% unstructured sparsity in Table \ref{tab:seq}. It's shown that our method consistently achieves better pruning performance, especially when the sequence length is relatively small, yielding much lower perplexity compared to both RIA and Wanda. This aligns with our theoretical analysis: with the decreasing of the input sequence length, our method has higher advantage over Wanda,
empirically supporting the theoretical derivation provided in the Appendix \ref{appb}. While RIA shows improved performance with longer sequences, it also incurs more expensive time costs. Therefore, our approach offers a balanced trade-off, maintaining both accuracy and time efficiency.

\subsection{Detailed Results on Zero-shot Accuracy} \label{appc3}

\begin{table}[htbp]
\centering
\vspace{-0.13in}
\caption{Accuracy comparison on 7 zero-shot tasks with OPT-6.7B}
\small
\vspace{0.05in}
\resizebox{0.99\columnwidth}{!}{
\begin{tabular}{c|c|c|ccccccc|c}
\toprule
\textbf{Model} & \textbf{Sparsity} & \textbf{Method} & \textbf{BoolQ} & \textbf{RTE} & \textbf{HellaSwag} & \textbf{WinoGrande} & \textbf{ARC-e} & \textbf{ARC-c} & \textbf{OBQA} & \textbf{Avg.} \\
\midrule
\multirow{9}{*}{OPT-6.7B} 
& \multirow{3}{*}{50\%} 
&  Wanda-16     & 62.32 & \textbf{53.43} & 46.03 & 61.48 & 62.66 & 27.22 & 24.60 & 48.24 \\
& & RIA-16       & 63.97 & 52.71 & 46.69 & \textbf{61.33} & 62.88 & 27.90 & 24.40 & 48.55 \\
& & CosP+VarP-16 (Ours) & \textbf{66.29} & 53.09 & \textbf{47.69} & 61.04 & \textbf{63.74} & \textbf{28.28} & \textbf{25.80} & \textbf{49.42} \\
\cmidrule{2-11}
& \multirow{3}{*}{2:4} 
& Wanda-16     & 62.17 & 52.35 & 40.91 & 59.74 & 56.06 & 24.74 & 21.60 & 45.36 \\
& & RIA-16       &   62.19    &   \textbf{53.42}    &   41.29    &   61.01    &   56.88    & 25.50      &   21.80    &   46.01    \\
& & CosP+VarP-16 (Ours) & \textbf{63.57} & 51.64 & \textbf{42.68} & \textbf{59.99} & \textbf{58.13} & \textbf{26.38} & \textbf{22.80} & \textbf{46.46} \\
\cmidrule{2-11}
& \multirow{3}{*}{4:8} 
& Wanda-16     & 62.22 & 53.41 & 43.45 & 60.13 & 58.95 & 26.71 & 23.00 & 46.83 \\
& & RIA-16       &   63.15    &   \textbf{53.79}    &   43.98    &   60.77    &   59.05    &   26.88    &   23.60    &   47.31    \\
& & CosP+VarP-16 (Ours) & \textbf{64.80} & 52.35 & \textbf{45.55} & \textbf{61.72} & \textbf{61.20} & \textbf{27.48} &\textbf{ 25.20} & \textbf{48.61} \\
\bottomrule
\end{tabular}
}
\label{tab:comp-opt6.7b}
\end{table}

\begin{table}[htbp]
\centering
\vspace{-0.16in}
\caption{Accuracy comparison on 7 zero-shot tasks with OPT-13B}
\small
\vspace{0.05in}
\resizebox{0.99\columnwidth}{!}{
\begin{tabular}{c|c|c|ccccccc|c}
\toprule
\textbf{Model} & \textbf{Sparsity} & \textbf{Method} & \textbf{BoolQ} & \textbf{RTE} & \textbf{HellaSwag} & \textbf{WinoGrande} & \textbf{ARC-e} & \textbf{ARC-c} & \textbf{OBQA} & \textbf{Avg.} \\
\midrule
\multirow{9}{*}{OPT-13B} 
& \multirow{3}{*}{50\%} 
&  Wanda-16     & 65.75 & 53.43 & 48.39 & 62.66 & 64.10 & 29.78 & 26.20 & 50.04 \\
& & RIA-16       & \textbf{65.86} & 54.11 & 49.01 & 62.84 & 64.31 & 30.90 & \textbf{26.60} &  \textbf{50.52} \\
& & CosP+VarP-16 (Ours) & 63.00 & \textbf{55.24} & \textbf{49.94} & \textbf{63.14} & \textbf{65.49} & \textbf{31.91} & 25.00 & \textbf{50.52}\\
\cmidrule{2-11}
& \multirow{3}{*}{2:4} 
& Wanda-16     & 64.31 & 52.70 & 43.80 & 60.93 & 59.55 & 26.11 & \textbf{22.60} & 47.14 \\
& & RIA-16       & \textbf{65.65} & 52.70 & 44.57 & 61.88 & 58.63 & 27.05 & 21.60 & 47.44 \\
& & CosP+VarP-16 (Ours) & 62.14 & \textbf{53.79} & \textbf{46.78} & \textbf{62.51} & \textbf{60.86} & \textbf{28.07} & \textbf{22.60} & \textbf{48.10} \\
\cmidrule{2-11}
& \multirow{3}{*}{4:8} 
& Wanda-16     & 65.29 & 53.07 & 46.16 & 62.66 & 60.81 & 27.30 & 24.80 & 48.58 \\
& & RIA-16       & \textbf{65.44} & 52.34 & 46.54 & \textbf{63.38} & 60.86 & 27.64 & 24.80 & 48.71 \\
& & CosP+VarP-16 (Ours) & 61.45 & \textbf{57.77} & \textbf{48.47} & 62.83 & \textbf{62.96} & \textbf{30.29} & \textbf{25.00} & \textbf{49.82} \\
\bottomrule
\end{tabular}
}
\label{tab:comp-opt13b}
\end{table}

\begin{table}[htbp]
\centering
\vspace{-0.16in}
\caption{Accuracy comparison on 7 zero-shot tasks with LLaMa-7B}
\small
\vspace{0.05in}
\resizebox{0.99\columnwidth}{!}{
\begin{tabular}{c|c|c|ccccccc|c}
\toprule
\textbf{Model} & \textbf{Sparsity} & \textbf{Method} & \textbf{BoolQ} & \textbf{RTE} & \textbf{HellaSwag} & \textbf{WinoGrande} & \textbf{ARC-e} & \textbf{ARC-c} & \textbf{OBQA} & \textbf{Avg.} \\
\midrule
\multirow{9}{*}{LLaMa-7B} 
& \multirow{3}{*}{50\%} 
&  Wanda-16     & 70.73 & 54.51 & 51.51 & 64.87 & 69.48 & \textbf{36.09} & \textbf{29.00} & 53.74 \\
& & RIA-16       & \textbf{71.54} & 61.73 & 51.49 & \textbf{66.65} & \textbf{69.75} & 35.90 & 28.40 & 55.07 \\
& & CosP+VarP-16 (Ours) & 70.67 & \textbf{64.62} & \textbf{51.61} & 66.61 & 69.07 & 35.58 & 27.60 & \textbf{55.11} \\
\cmidrule{2-11}
& \multirow{3}{*}{2:4} 
& Wanda-16     & 68.19 & 53.79 & 41.73 & 62.04 & 59.80 & 26.70 & 22.60 & 47.83 \\
& & RIA-16       &   67.98    &    55.23   &    42.03   &   62.03    &   60.48    &   26.96    &   \textbf{24.20}   &   48.41    \\
& & CosP+VarP-16 (Ours) & \textbf{68.87} & \textbf{56.68} & \textbf{43.88} & \textbf{63.06} & \textbf{61.74} & \textbf{28.59} & 22.80 & \textbf{49.37} \\
\cmidrule{2-11}
& \multirow{3}{*}{4:8} 
& Wanda-16     & \textbf{70.00} & 55.23 & 46.81 & 64.09 & 63.38 & 31.91 & 24.80 & 50.88 \\
& & RIA-16       &   69.29    &   55.95    &   47.00   &   \textbf{64.48}   &   63.72    &   31.65    &   \textbf{26.20}   &   51.18   \\
& & CosP+VarP-16 (Ours) & 69.85 & \textbf{59.93} & \textbf{48.42} & 64.25 & \textbf{64.02} & \textbf{32.34} & 25.60 & \textbf{52.05} \\
\bottomrule
\end{tabular}
}
\label{tab:comp-llama7b}
\end{table}

\begin{table}[htbp]
\centering
\vspace{-0.16in}
\caption{Accuracy comparison on 7 zero-shot tasks with LLaMa-13B}
\small
\vspace{0.05in}
\resizebox{0.99\columnwidth}{!}{
\begin{tabular}{c|c|c|ccccccc|c}
\toprule
\textbf{Model} & \textbf{Sparsity} & \textbf{Method} & \textbf{BoolQ} & \textbf{RTE} & \textbf{HellaSwag} & \textbf{WinoGrande} & \textbf{ARC-e} & \textbf{ARC-c} & \textbf{OBQA} & \textbf{Avg.} \\
\midrule
\multirow{9}{*}{LLaMa-13B} 
& \multirow{3}{*}{50\%} 
&  Wanda-16     & 73.35 & 58.12 & 55.12 & 70.51 & \textbf{74.17} & 41.21 & \textbf{31.20} & \textbf{57.67} \\
& & RIA-16       & 73.34 & 57.11 & 54.51 & 70.62 & 73.94 & 40.87 & 30.60 & 57.28 \\
& & CosP+VarP-16 (Ours) & \textbf{73.98} & \textbf{58.48} & \textbf{54.97} & \textbf{70.64} & 72.65 & \textbf{41.47} & 29.00 & 57.31 \\
\cmidrule{2-11}
& \multirow{3}{*}{2:4} 
& Wanda-16     & 70.12 & \textbf{53.79} & 46.44 & 66.45 & 65.82 & 32.25 & 25.80 & 51.52 \\
& & RIA-16       &   69.85    &   53.42   &   47.43    &   \textbf{67.30}  &   \textbf{66.92}   &   33.68   &   \textbf{26.20}   &   52.11   \\
& & CosP+VarP-16 (Ours) & \textbf{70.59} & 53.14 & \textbf{49.16} & 66.61 & 66.49 & \textbf{33.94} & \textbf{26.20} & \textbf{52.30} \\
\cmidrule{2-11}
& \multirow{3}{*}{4:8} 
& Wanda-16     & 70.69 & \textbf{54.15} & 50.73 & 68.67 & 70.17 & 37.82 & 27.60 & 54.26 \\
& & RIA-16       &   71.16   &   53.43    &  51.24   &  \textbf{70.48}   &  69.95   &   37.37   &  28.00   &  54.52  \\
& & CosP+VarP-16 (Ours) & \textbf{72.68} & 53.09 & \textbf{52.57} & 67.62 & \textbf{71.07} & \textbf{38.99} & \textbf{28.80} & \textbf{54.97} \\
\bottomrule
\end{tabular}
}
\label{tab:comp-llama13b}
\end{table}

\begin{table}[htbp]
\centering
\vspace{-0.16in}
\caption{Accuracy comparison on 7 zero-shot tasks with LLaMa-2-7B}
\small
\vspace{0.05in}
\resizebox{0.99\columnwidth}{!}{
\begin{tabular}{c|c|c|ccccccc|c}
\toprule
\textbf{Model} & \textbf{Sparsity} & \textbf{Method} & \textbf{BoolQ} & \textbf{RTE} & \textbf{HellaSwag} & \textbf{WinoGrande} & \textbf{ARC-e} & \textbf{ARC-c} & \textbf{OBQA} & \textbf{Avg.} \\
\midrule
\multirow{9}{*}{LLaMa-2-7B} 
& \multirow{3}{*}{50\%} 
&  Wanda-16     & 74.44 & 55.59 & 51.29 & 66.12 & \textbf{71.63} & 37.25 & \textbf{29.40} & \textbf{55.10} \\
& & RIA-16       & 73.70 & 55.59 & 51.37 & \textbf{66.45} & 69.69 & 35.66 & 28.60 & 54.43 \\
& & CosP+VarP-16 (Ours) & \textbf{73.72} & \textbf{55.62} & \textbf{51.53} & 66.08 & 70.48 & \textbf{37.47} & 29.20 & 54.87  \\
\cmidrule{2-11}
& \multirow{3}{*}{2:4} 
& Wanda-16     & \textbf{67.83} & 53.43 & 39.85 & 59.59 & 59.30 & 27.04 & \textbf{21.80} & 46.98 \\
& & RIA-16       &    66.90   &   53.39   &   40.27    &  59.59   &   60.77   &  27.98    &   \textbf{21.80}   &   47.24   \\
& & CosP+VarP-16 (Ours) & 65.34 & \textbf{54.20} & \textbf{42.45} & \textbf{61.29} & \textbf{62.54} & \textbf{28.62} & 20.80 & \textbf{47.89} \\
\cmidrule{2-11}
& \multirow{3}{*}{4:8} 
& Wanda-16     & 71.56 & 54.15 & 45.29 & 63.69 & \textbf{65.48} & 32.51 & \textbf{24.80} & 51.07 \\
& & RIA-16       &   \textbf{72.88}   &   54.14    &  45.82   &   64.15  &  66.27   &   32.75   &  \textbf{24.80}   &  51.54  \\
& & CosP+VarP-16 (Ours) & 69.70 & \textbf{54.51} & \textbf{47.30} & \textbf{64.72} & 66.37 & \textbf{34.56} & 24.60 & \textbf{51.68} \\
\bottomrule
\end{tabular}
}
\label{tab:comp-llama2-7b}
\end{table}

\begin{table}[htbp]
\centering
\vspace{-0.16in}
\caption{Accuracy comparison on 7 zero-shot tasks with LLaMa-2-13B}
\small
\vspace{0.05in}
\resizebox{0.99\columnwidth}{!}{
\begin{tabular}{c|c|c|ccccccc|c}
\toprule
\textbf{Model} & \textbf{Sparsity} & \textbf{Method} & \textbf{BoolQ} & \textbf{RTE} & \textbf{HellaSwag} & \textbf{WinoGrande} & \textbf{ARC-e} & \textbf{ARC-c} & \textbf{OBQA} & \textbf{Avg.} \\
\midrule
\multirow{9}{*}{LLaMa-2-13B} 
& \multirow{3}{*}{50\%} 
&  Wanda-16     & 78.66 & \textbf{62.45} & 56.61 & \textbf{70.56} & 76.72 & \textbf{42.58} & 32.00 & 59.94 \\
& & RIA-16       & 79.90 & 62.09 & 56.31 & 70.00 & 76.18 & 40.27 & 31.00 & 59.39 \\
& & CosP+VarP-16 (Ours) & \textbf{80.57} & 60.94 & \textbf{57.49} & 70.16 & \textbf{76.94} & 42.49 & \textbf{32.20} & \textbf{60.11} \\
\cmidrule{2-11}
& \multirow{3}{*}{2:4} 
& Wanda-16     & \textbf{77.42} & \textbf{59.55} & 46.01 & \textbf{66.92} & 68.57 & 33.10 & 23.60 & 53.59 \\
& & RIA-16       &   77.18    &   58.84   &   46.83    &  66.69   &   68.77   &  33.27    &   23.40   &  53.57    \\
& & CosP+VarP-16 (Ours) & 76.97 & 58.42 & \textbf{49.07} & 64.03 & \textbf{69.79} & \textbf{34.97} & \textbf{25.20} & \textbf{54.06} \\
\cmidrule{2-11}
& \multirow{3}{*}{4:8} 
& Wanda-16     & 79.35 & 60.29 & 51.67 & 68.03 & \textbf{74.03} & \textbf{38.90} & 27.80 & 57.15 \\
& & RIA-16       &  \textbf{79.75}    &   60.29    &  51.85   &   \textbf{68.51}  &  73.10   &  38.31    &  27.00   &  56.97  \\
& & CosP+VarP-16 (Ours) & 77.49 & \textbf{62.77} & \textbf{53.63} & 67.94 & 73.18 & 37.60 & \textbf{29.20} & \textbf{57.40}  \\
\bottomrule
\end{tabular}
}
\label{tab:comp-llama2-13b}
\end{table}

To further evaluate the performance of our proposed CosP+VarP method, 
we conduct a comparison against 
baseline methods (i.e., Wanda and RIA), across seven few-shot tasks, i.e.,  BoolQ~\cite{clark2019boolq}, RTE~\cite{wang2018glue}, HellaSwag~\cite{zellers2019hellaswag}, WinoGrande~\cite{sakaguchi2019winogrande}, ARC-Easy, ARC-Challenge~\cite{clark2018arc}, and OpenBookQA~\cite{mihaylov2018openbookqa}. 
The experimental results on 6 large language models (i.e., OPT-6.7B, OPT-13B, LLaMa-7B, LLaMa-13B, LLaMa-2-7B, LLaMa-2-13B) are shown from Table \ref{tab:comp-opt6.7b} to Table \ref{tab:comp-llama2-13b}.
The best results are marked in bold. CosP+VarP-16 denotes our proposed method CosP+VarP with an input sequence length of 16. Experimental results show that CosP+VarP-16 has best accuracy on most of the tasks among three different sparsity patterns with all different models, demonstrating the effectiveness of our proposed method.




\end{document}